%% file: ms.tex
\documentclass[twoside]{article}
\usepackage[accepted]{aistats2021}
\usepackage[round]{natbib}

\usepackage[shortlabels]{enumitem}
\usepackage[ruled,vlined]{algorithm2e}
\usepackage{caption}
\usepackage{subcaption}
\input{defs}

\begin{document}

%

%
\runningauthor{Wu$^*$, Miller$^*$, Anderson, Pleiss, Blei, Cunningham}

\twocolumn[

\aistatstitle{Hierarchical Inducing Point Gaussian Process for Inter-domain Observations}
\aistatsauthor{ Luhuan Wu$^{*1}$ \And Andrew Miller$^{*1}$ \And  Lauren Anderson$^2$ \And Geoff Pleiss$^1$}
\vspace{0.2cm}
\aistatsauthor{David Blei$^1$ \And John Cunningham$^1$}

\aistatsaddress{ $^1$ Columbia University \And  $^2$ The Observatories of the Carnegie Institution for Science \\ \{lw2827, gmp2162, david.blei, jpc2181\}@columbia.edu   \qquad \qquad \qquad \qquad \qquad \qquad  \qquad \qquad \qquad
\\  \{andrew.colin.miller, anders.astro\}@gmail.com
\qquad \qquad \qquad \qquad \qquad \qquad  \qquad \qquad \qquad } ]

\begin{abstract}
\input{sections/abstract.tex}
\end{abstract}

\input{sections/intro.tex}

\input{sections/background.tex}

\input{sections/method.tex}

\input{sections/related-work.tex}

\input{sections/experiments.tex}

\input{sections/discussion.tex}

\input{sections/acknowledgement.tex}
\bibliographystyle{apalike}
\bibliography{refs.bib}

\clearpage
\appendix

\makeatletter
	\setcounter{table}{0}
	\renewcommand{\thetable}{S\arabic{table}}%
	\setcounter{figure}{0}
	\renewcommand{\thefigure}{S\arabic{figure}}%
    \setcounter{equation}{0}
    \renewcommand\theequation{S\arabic{equation}}
\makeatother

\input{appendix-arxiv.tex}

\end{document}

%% file: defs.tex
\usepackage[usenames,dvipsnames]{xcolor}

\renewcommand{\paragraph}[1]{\textbf{#1}~}

\usepackage[colorlinks,linktoc=all]{hyperref}
\usepackage[all]{hypcap}
\hypersetup{citecolor=MidnightBlue}
\hypersetup{linkcolor=MidnightBlue}
\hypersetup{urlcolor=MidnightBlue}

\usepackage{booktabs}       
\usepackage{multirow}

\usepackage{enumitem}

\usepackage[makeroom]{cancel}
\usepackage{amsmath, amssymb}
\usepackage{amsfonts}       
\usepackage[nice]{nicefrac}
\usepackage{tikz}
\newcommand\circled[1]{%
  \tikz[baseline=(X.base)] 
    \node (X) [draw, shape=circle, inner sep=0] {\strut #1};}

\newcommand{\given}{\,|\,}

\newcommand{\ba}{\boldsymbol{a}}
\newcommand{\bb}{\boldsymbol{b}}
\newcommand{\bc}{\boldsymbol{c}}

\newcommand{\be}{\boldsymbol{e}}

\newcommand{\bk}{\boldsymbol{k}}

\newcommand{\bm}{\boldsymbol{m}}

\newcommand{\bp}{\boldsymbol{p}}

\newcommand{\br}{\boldsymbol{r}}
\newcommand{\bs}{\boldsymbol{s}}

\newcommand{\bu}{\boldsymbol{u}}
\newcommand{\bv}{\boldsymbol{v}}

\newcommand{\bx}{\boldsymbol{x}}
\newcommand{\by}{\boldsymbol{y}}
\newcommand{\bz}{\boldsymbol{z}}

\newcommand{\bA}{\boldsymbol{A}}
\newcommand{\bB}{\boldsymbol{B}}
\newcommand{\bC}{\boldsymbol{C}}
\newcommand{\bD}{\boldsymbol{D}}

\newcommand{\bF}{\boldsymbol{F}}

\newcommand{\bI}{\boldsymbol{I}}

\newcommand{\bK}{\boldsymbol{K}}
\newcommand{\bL}{\boldsymbol{L}}

\newcommand{\bP}{\boldsymbol{P}}

\newcommand{\bR}{\boldsymbol{R}}
\newcommand{\bS}{\boldsymbol{S}}
\newcommand{\bT}{\boldsymbol{T}}

\newcommand{\blambda}{{\boldsymbol{\lambda}}}
\newcommand{\btheta}{{\boldsymbol{\theta}}}

\newcommand{\bLambda}{{\boldsymbol{\Lambda}}}

\newcommand{\brho}{\boldsymbol{\rho}}
\newcommand{\etab}{\boldsymbol{\eta}}
\newcommand{\bepsilon}{\boldsymbol{\epsilon}}

%% file: sections/abstract.tex
\noindent

We examine the general problem of inter-domain Gaussian Processes (GPs):
problems where the GP realization and
the noisy observations of that realization lie on different domains.
When the mapping between those domains is linear,
such as integration or differentiation,
inference is still closed form.
However, many of the scaling and approximation techniques
that our community has developed do not apply to this setting.
In this work, we introduce
the \emph{hierarchical inducing point GP} (HIP-GP),
a scalable inter-domain GP inference method
that enables us to improve the approximation accuracy
by increasing the number of inducing points to the millions.
HIP-GP, which
relies on inducing points with grid structure and a stationary kernel assumption,
is suitable for low-dimensional problems.
In developing HIP-GP, we introduce (1) a fast whitening strategy,
and (2) a novel preconditioner for conjugate gradients which
can be helpful in general GP settings. 
Our code is available at \url{https://github.com/cunningham-lab/hipgp}. 

%% file: sections/intro.tex
\section{INTRODUCTION}
\label{sec:intro}
Gaussian processes (GPs) are a widely used statistical tool for inferring
unobserved functions \citep{cressie1992statistics, cressie1990origins,
rasmussen2006gaussian}. The classic goal of GPs is to infer the
unknown function given noisy observations. Here, we are interested in a more
general setting, inter-domain GPs,
where the observed data is related to the latent function via some linear transformation,
such as integration or differentiation,
while an identity transformation recovers the standard GP problem.
One motivating example is an
astrophysics problem: mapping the three-dimensional spatial distribution of dust in the Milky way \citep{green2015three,
leike2019charting, kh2017inferring}.
Interstellar dust is a latent function that can be inferred from star observations. However, because we are embedded in our own dust field,
we can only observe some noisy \textit{integral} of the dust function along the line of sight between Earth and a star.
Beyond this example, inter-domain GPs arise often in the literature: integrated observations have been used in probabilistic construction of optimization algorithms \citep{wills2017construction}, quadrature rules \citep{minka2000deriving}, and tomographic reconstructions \citep{jidling2018probabilistic};
while derivative observations have been used in dynamical systems \citep{solak2003derivative}, 
modeling monotonic functions \citep{riihimaki2010gaussian} and Bayesian optimization \citep{garnett2010bayesian, siivola2018correcting}.

In practice, this type of inter-domain GP problem poses two interwined obstacles that are beyond the reach of current techniques.
First, large-scale exact modeling is usually intractable.
The joint distribution of inter-domain observations and the underlying GP involves 
the transformed-domain and inter-domain kernel expressions,
which rarely admits analytical solutions
and requires approximations \citep{lazaro2009inter, hendriks2018evaluating}.
Common approximations are often handled
by numerical integration, which is infeasible for big datasets since it requires
integrating all pairwise correlations.

Moreover, inter-domain GPs suffer from the
same scalability issues as regular GPs.
For a dataset with $N$ observations,
the likelihood function depends on $N^2$ pairwise correlations.
The leading strategy to scale standard GP inference is to use $M \ll N$ \emph{inducing points}
to represent the global behavior of GP functions \citep{rasmussen2006gaussian}.
One popular inducing point method is stochastic variational Gaussian process (SVGP),
which factorizes the objective over mini-batches of data
and requires only $O(M^2)$ storage and $O(M^3)$ computation \citep{hensman2013gaussian}.
In the current practice of SVGP, $M$ is limited to under $10{,}000$ \citep{wilson2015kernel, izmailov2018scalable}.
However, many inter-domain problems are spatial or temporal in nature,
and the data do not lie in some small manifold in that space.
In the interstellar dust problem, for example, we aim to make inference at
every point in a dense 3D space.
A small set of inducing points
is incapable of resolving the resolution of interest,
which is around 4 orders of magnitude smaller than the domain size.
Furthermore, \citet{bauer2016understanding} shows that more inducing points are needed
to reduce the overestimated observation noise parameter induced by SVGP.
All of these facts necessitate the need to scale both $N$ and $M$ to larger quantities.

To this end, we develop the hierarchical inducing point GP (HIP-GP),
a method to scale GP inference
to millions of inducing points and observations for spatial-temporal inter-domain problems.
In particular,
\begin{itemize}
  \item We adapt the SVGP framework to inter-domain settings
  by decoupling observations and inducing points into different domains.
  This framework alleviates the difficulties of computing the full transformed kernel matrices,
  and enables the exploitations of the latent kernel structure.
  \item We then develop the HIP-GP algorithm to address the computational bottlenecks
   of standard SVGP objectives, employing two core strategies:
   \begin{itemize}
     \item  Fast matrix inversion with conjugate gradient method
     using the \emph{hierachical Toeplitz structure}.
      Upon this structure, we design a novel \emph{preconditioner}
     and a new \emph{whitening strategy} to further speed up computations;
    \item A \emph{structured variational approximation} of
    the posterior over inducing point values.
  \end{itemize}
\end{itemize}
HIP-GP is suitable for low-dimensional inter-domain GP problems,  and applies in
settings where the kernel function is stationary and inducing points fall on a fixed, evenly-spaced grid.
In addition, the technical innovations in developing HIP-GP
are useful in a variety of more general settings.



%% file: sections/background.tex
\section{BACKGROUND}
\label{sec:background}

\subsection{Inter-domain GPs}

Following the notations in \citet{van2020framework},
we consider a statistical model of the form
\begin{align}
  \rho &\sim GP\left( 0, k_{\theta}(\cdot, \cdot) \right) \\
  \rho^* &= \mathcal{L} \circ \rho \\
  y_n \given \bx_n, \rho^* &\sim \mathcal{N}(\rho^*(\bx_n), \sigma_n^2)
\end{align}
for a dataset of $N$ observations $\mathcal{D} \triangleq \{ y_n, \bx_n,
\sigma_n^2 \}_{n=1}^N$, where $\mathcal{L}$ is a linear operator
and $k_{\theta}(\cdot, \cdot)$ is the covariance function that encodes prior assumptions
about the function $\rho$. 
Note that GPs are closed under linear operators,
therefore $\rho^*$ is also a GP \citep{rasmussen2006gaussian}.

One common linear operator is the integral operator,
$\mathcal{L} \circ \rho (\cdot)  = \int \rho( \bx) w (\bx) d \bx$,
as used in \citet{lazaro2009inter}.
We see that this $\mathcal{L}$ maps the entire function
$\rho (\cdot)$ to a single real value.
Another example is the derivative of the $d$th input dimension
$\mathcal{L} \circ \rho (\cdot) = \frac{ \partial \rho }{\partial x_d } (\bx_n)$.
In this case, the operator only depends on the neighborhood around $\bx_n$.
Derivative observations are often useful for algorithmic purposes, e.g. in \citet{riihimaki2010gaussian}.
In application problems, they
could be either collected, e.g.  velocity measured by physical detectors,
or identified from function observations \citep{solak2003derivative}.
We also notice that setting $\mathcal{L}$ to an identity map fits
 regular GPs into this framework.

The goal of inter-domain GPs is to infer the underlying
function $\rho(\bx)$ --- either to compute $p(\rho(\bx_*) \given \mathcal{D})$
for new test locations $\bx_*$ or to improve estimates of $p(\rho(\bx_n) \given
\mathcal{D})$ for observed location $\bx_n$ given \emph{all} observations.

\subsection{Stochastic Variational Gaussian Process}
\label{sec:SVGP}
The stochastic variational Gaussian process (SVGP) is an approximate method that
scales GP inference to large $N$ \citep{hensman2013gaussian}.
Denote the $M$ inducing point \emph{locations} $\bar{\bx} = \left( \bar{\bx}_1, \dots,
\bar{\bx}_M \right)$, and the vector of inducing point \emph{values}
$\bu \triangleq \left( \rho(\bar{\bx}_1), \dots, \rho(\bar{\bx}_M) \right)$.
SVGP defines a variational distribution over the inducing point 
values $\bu$ and the latent process values
$\brho \triangleq \left(\rho(\bx_1), \dots, \rho(\bx_N) \right)$ of the form
\begin{align*}
    q(\bu, \brho) = q_{\blambda}(\bu) p(\brho \given \bu) \, , \quad
    q_{\blambda}(\bu) = \mathcal{N}(\bu \given \bm, \bS) \,,
\end{align*}
where $q_\blambda(\bu)$ is a multivariate Gaussian,
$p(\brho \given \bu)$ is determined by the GP prior
and $\blambda \triangleq (\bm, \bS)$ are variational parameters.
This choice of variational family induces a convenient cancellation, resulting
in a separable objective
\citep{titsias2009variational}
\begin{align}
&\mathcal{L}(\blambda) \label{eq:svgp-elbo} \\
  &= \underbrace{\mathbb{E}_{q_{\blambda}(\bu)}\left[
      \mathbb{E}_{p(\brho \given \bu)} \left[ \ln p(\by \given \brho) \right]
    \right]}_{\circled{a}} -
    \underbrace{KL(q_{\blambda}(\bu)\,||\,p(\bu))}_{\circled{b}} \nonumber\,.
\end{align}
We can write \circled{a} as a sum over $N$ observations
\begin{align}
    \circled{a}
    &= \sum_{n=1}^N \underbrace{
          \mathbb{E}_{q_{\blambda}(\bu)}\left[
              \mathbb{E}_{p(\rho_n \given \bu)} \left[
                  \ln p(y_n \given \rho_n)
              \right]
          \right]
        }_{\triangleq \circled{$a_n$}} \,.
\end{align}
The factorization of \circled{$a_n$}
enables the objectives to be estimated with mini-batches in a large dataset.
However, notice that \circled{$b'$}, the KL-divergence of two Gaussians,
will involve a term $\ln |\bK_{\bu,\bu}|$
which requires $O(M^3)$ computation.

\subsection{Matrix Solves with Conjugate Gradients}
Conjugate gradients (CG) is an iterative algorithm for solving a linear
system using only matrix-vector multiplies (MVM).
CG computes $\bK^{-1} \bp$ for any $\bp \in \mathbb{R}^M$ by computing
$\bK \bv$ for a sequence of vectors $\bv \in \mathbb{R}^M$ determined by the algorithm.
For $\bK$ of size $M \times M$, CG computes the exact solution after $M$
iterations,
and typically converges after some smaller number of steps $S < M$
\citep{hestenes1952methods, nocedal2006numerical}.

Preconditioned conjugate gradients (PCG) is an augmented version of CG that
solves the system in a transformed space.  A good preconditioner can
dramatically speed up convergence \citep{shewchuk1994introduction,
cutajar2016preconditioning}.

%% file: sections/method.tex
\section{SCALING $M$: HIP-GP for INTER-DOMAIN PROBLEMS}
\label{sec:hip-gps}
We first formulate the SVGP framework for inter-domain observations, and identify
its computational bottlenecks in Section~\ref{sec:method-interdomain-svgp}.
We then address these bottlenecks by the HIP-GP algorithm using the techniques developed in
Section~\ref{sec:computational} - \ref{sec:variational-families}.
In Section~\ref{sec:method-summary}, we summarize our methods and discuss
optimization procedures for HIP-GP.
\input{sections/method-inter-domain-svgp.tex}

\input{sections/method-computation.tex}

\input{sections/method-variational-approx.tex}
\subsection{Method Summary}
\label{sec:method-summary}
The modeling difficulty of inter-domain GP problems arises from the numerical intractability
of computing the full transformed-domain covariance
$\bK^{**}_{N,N}$ of size $N \times N$. We avoid this difficulty by decoupling the
 the observations and the inducing points into different domains under the SVGP framework.
Moreover, we leverage the kernel structure of the Gram matrix
$\bK^{}_{\bu,\bu}$ in the latent domain for efficient computations.

The computational
difficulty stems from the computations with the kernel matrix $\bK_{\bu,\bu}$ and the variational covariance $\bS$.  We avoid having to
compute $\ln|\bK^{}_{\bu,\bu}|$ by using a \emph{whitened parameterization};
we develop a \emph{fast whitening strategy} to compute the whitened correlation term
 $\bk_n = \bR^\top \bK_{\bu, \bu}^{-1} \bk^{*}_{\bu,n}$
by exploiting  the \emph{hierarchical Toeplitz structure} with a \emph{novel
preconditioner};
and finally we explore a \emph{structred representation} for $\bS$.

\paragraph{Optimization} We perform natural gradient descent on
variational parameters using closed-form gradient updates.
For gradient-based learning of kernel hyperparameters, automatically differentiating
through the CG procedure is not numerically stable.
Fortunately, we can efficiently compute the analytical gradient of CG solves 
utilizing the hierarchical Toeplitz structure,
without increasing the computational complexity.
See appendix for more details on gradient derivations.

%% file: sections/method-inter-domain-svgp.tex
\subsection{Inter-domain SVGP Formulation}
\label{sec:method-interdomain-svgp}
We show that the inter-domain observations can be easily incorporated into the SVGP framework.
We place a set of inducing points $\bu = \rho(\bar{\bx})$ in the latent domain
at input locations $\bar{\bx} = \left( \bar{\bx}_1, \cdots, \bar{\bx}_M \right)$.
Connections to the observations are made through the
inter-domain covariance,
while the observations are characterized by
the transformed-domain covariance.
Formally, we have the inter-domain GP prior:
\begin{align}
  \label{eq:inter-domain-GP}
\begin{pmatrix}
\rho_n^* \\
\bu
\end{pmatrix}
\sim \mathcal{N} \left( 0,
\begin{pmatrix}
 k^{**}_{n,n} & \bk^*_{n, \bu} \\
 \bk^*_{\bu, n} & \bK_{\bu, \bu} \\
\end{pmatrix}
\right),
\end{align}
where the inter-domain covariance and the transformed-domain covariance are defined as
\begin{align}
\label{eq:domain-kernel}
\bk^*_{\bu, n} &\triangleq Cov \left(\rho(\bar{\bx}), \rho^*(\bx_n)\right)
 = Cov \left(\bu, \rho^*_n \right)  \, , \\
\label{eq:domain-kernel2}
k^{**}_{n,n} &\triangleq Cov \left(\rho^*(\bx_n), \rho^*(\bx_n) \right)
 = Cov \left( \rho^*_n, \rho^*_n  \right) \, , 
\end{align}
and the latent domain covariance is
\begin{align}
  \bK_{\bu, \bu} &\triangleq Cov \left( \rho (\bar{\bx}), \rho (\bar{\bx}) \right)
    = Cov \left( \bu, \bu \right).
\end{align}
This form of the prior suggests formulating the inter-domain SVGP objective
as follows
\begin{align}
\mathcal{L}(\blambda) \label{eq:inter-domain-svgp} &= \sum_{n=1}^N \mathbb{E}_{q_{\blambda}(\bu)}\left[
      \mathbb{E}_{p(\rho_n^* \given \bu)} \left[ \ln p(y_n \given \rho_n^*) \right] \right] \\
& \qquad  - KL(q_{\blambda}(\bu)\,||\,p(\bu)) \nonumber\,,
\end{align}
where
\begin{align*}
  p(\rho_n^* | \bu) &= N(\rho_n^* | \bk_{n,\bu}^* \bK_{\bu, \bu}^{-1} \bu,
  k_{n,n}^{**} - \bk_{n,\bu}^* \bK_{\bu, \bu}^{-1} \bk_{\bu, n}^*)\,.
\end{align*}
Note that this framework can be extended to observations in multiple domains
by including them with their corresponding inter-domain and transformed-domain covariances.
Under this formulation, we avoid computing the $N \times N$
transformed-domain covariance matrix $\bK_{N,N}^{**}$ that appears in the exact GP objective.
Instead, only $N$ terms of variance $k_{n,n}^{**}$ need to be evaluated. 
Importantly, the disentanglement of observed and latent domains
enables us to exploit structure of  $\bK_{\bu, \bu}$ for efficient computations.
Such exploitation would be difficult without a variational approximation,
especially in the case of mixed observations from multiple domains.

\paragraph{Whitened Parameterization}
Whitened parameterizations are used to improve
inference in models with correlated priors because they offer a
better-conditioned posterior
\citep{murray2010slice, hensman2015mcmc}.
Here we will show an additional computational benefit
in the variational setting --- the whitened posterior allows us to
avoid computing $\ln |\bK_{\bu,\bu}|$ which appears in the KL term
in Equation~\ref{eq:inter-domain-svgp}.
To define the whitened parameterization, we describe the GP prior over
$\bu$ as a deterministic function of standard normal parameters $\bepsilon$: 
\begin{align}
  \bepsilon \sim \mathcal{N}(0, I) \,, \quad
  \bu = \bR \bepsilon.
\end{align}
To preserve the covariance structure in the prior distribution of $(\rho^*_n, \bu)$ (Equation~\ref{eq:inter-domain-GP}),
the transformation $\bR$ and
the whitened correlation $\bk_n \triangleq Cov(\bepsilon, \rho^*_n)$ need to satisfy the following two equalities:  
\begin{equation}
  \begin{aligned}
  \label{eq:whitened-criteria}
  \bK_{\bu, \bu} &=  Cov( \bR \bepsilon, \bR \bepsilon) =   \bR \bR^\top \, , \\
  \quad  \bk^{*}_{\bu, n} &= Cov(\bR \bepsilon, \rho^*_n) = \bR \bk_n \, . 
\end{aligned}
\end{equation}
The classical whitening strategy in GP inference is to use the Cholesky decomposition: $\bK_{\bu, \bu} = \bL \bL^\top$ where  $\bL$ is a lower triangular matrix. In this case, $\bR = \bL$ 
and $\bk_n = \bL^{-1} \bk^*_{\bu, n}$.

Now we can target the variational posterior over
the whitened parameters
$\bepsilon$: 
${q_{\blambda}(\bepsilon) = \mathcal{N}(\bepsilon \given \bm, \bS)}$.
The resulting \emph{whitened variational objective} is
\begin{align}
\label{eq:whitened-elbo}
\mathcal{L}(\blambda)
  &= \sum_{n} \underbrace{
  	    \mathbb{E}_{q_{\blambda}(\bepsilon) p(\rho_n^* \given \bepsilon)} \left[
	    	\ln p(y_n \given \rho_n^*)
	    \right]
	  }_{\circled{$a'_n$}} -
      \underbrace{
	  	KL(q_{\blambda}(\bepsilon) \,||\, p(\bepsilon))
	  }_{\circled{b$'$}} \,
\end{align}
where
\begin{align}
\circled{$a'_n$}
&= -\frac{1}{2}\ln \sigma_n^2 - \frac{1}{2\sigma_n^2}
    \Big( y_n^2 + k^{**}_{n,n} - \bk_n^\intercal \bk^{}_n  \nonumber \\
&\quad + \bk_n^\intercal
     \left( \bS + \bm \bm^\intercal \right) \bk_n - 2 y^{}_n \bk_n^\intercal \bm
    \Big) \,\,,\\
\circled{b$'$}
  &= \frac{1}{2}\left( \text{tr}(\bS) + \bm^\intercal \bm - \ln |\bS| - M \right) .
\label{eq:kl-term}
\end{align}
\paragraph{Computational Bottlenecks}
The whitened objective above still factorizes over data points. However, there remain two computational bottlenecks. First, the correlation term $\bk_n$ in \circled{$a_n'$}
depends on the choice of the whitening strategy.
The common Cholesky strategy requires $O(M^3)$ computation and $O(M^2)$ storage
which is infeasible for large $M$.
We address this bottleneck in Section~\ref{sec:computational}.
The second bottleneck lies in the variational covariance $\bS$ which is an $M \times M$ matrix,
requiring $O(M^2)$ to store and $O(M^3)$ to compute the $\ln |\bS|$ in \circled{$b'$}.
We will address this problem by a structured
variational approximation in Section~\ref{sec:variational-families}.

%% file: sections/method-computation.tex
\subsection{Computational Accelerations}
\label{sec:computational}
We now turn to the first bottleneck --- how to design an efficient whitening strategy
to compute the term $\bk_n$. 
To do so,
we rely on judicious placement of inducing points and assume a stationary
covariance function, a general and commonly used class.
We describe three key ingredients below.

\paragraph{Hierarchical Toeplitz Structure}
\label{sec:toeplitz-structure}
Consider a $D$-dimensional grid of evenly spaced points
of size $M \triangleq M_1 \times \cdots \times M_D$, characterized by
one-dimensional grids of size $M_i$ along dimension $i, i=1:D$,
where $D$ is the input dimension.
Under a stationary kernel, we construct a covariance matrix
for this set of points in $x$-major
order (i.e.~\texttt{C}-order).
Such a matrix will have \emph{hierarchical Toeplitz structure},
which means the diagonals of the matrix are constant.
Because of this data redundancy, a hierarchical Toeplitz matrix
is characterized by its first row.
Now we place the inducing points along
a fixed, equally-spaced grid, resulting in a $M \times M$ hierarchical
Toeplitz Gram matrix $\bK^{}_{\bu,\bu}$.
The efficient manipulation of $\bK_{\bu, \bu}$ is through its circulant embedding:
\begin{align}
    \bC =
    \begin{pmatrix}
    \bK_{\bu, \bu} & \tilde{\bK} \\
    \tilde{\bK}^\top & \bK_{\bu, \bu}
    \end{pmatrix}
\end{align}
where $\tilde{\bK}$ is the appropriate reversal of $\bK_{\bu, \bu}$ to make $\bC$ circulant.
$\bC$ admits a convenient diagonalization
\begin{align}
  \bC = \bF^\top \bD \bF
      = \bF^\top \text{diag} \left(\bF \bc \right) \bF \,\,,
\end{align}
where $\bF$ is the fast Fourier transform matrix, $\bD$ is a
diagonal matrix of $\bC$'s eigenvalues, and $\bc$ is the first row of $\bC$.
This diagonalization enables fast MVMs with $\bC$ and
hence the embedded $\bK_{\bu, \bu}$ via the FFT algorithm
in $O(M \ln M)$ time, further making it sufficient for use
within CG to efficiently solve a linear system.

The fast solves afforded by Toeplitz structure have been previously
utilized for exact GP inference \citep{cunningham2008fast, wilson2015thoughts}.
Here, we extend the applicability of Toeplitz structure
to the variational inter-domain case
by introducing a fast whitening procedure and an effective preconditioner for CG.

\paragraph{Fast Whitening Strategy}
\label{sec:toeplitz-whitening}
Similar to the Cholesky decomposition, we aim to
find a whitened matrix $\bR$ that serves as a root of $\bK_{\bu, \bu}$,
i.e. $\bR \bR^T = \bK_{\bu, \bu}$.
Directly solving $\bK_{\bu, \bu}^{1/2}$ is not trivial. Alternatively, we access the root from the circulant embedding of $\bK_{\bu, \bu}$.  We consider the square root of the circulant matrix
\begin{equation}\label{eq:csquared-fft}
  \bC^{1/2} = \bF^\intercal \bD^{1/2} \bF,
\end{equation}
and its block representation
\begin{equation}
  \bC^{1/2} = \begin{pmatrix}
  \bA & \bB \\
  \bB^\top & \bD \\
\end{pmatrix}.
\end{equation}
We make a key observation that the first row-block $(\bA, \bB)$ can be viewed as a
``rectangular root" of $\bK_{\bu, \bu}$. That is, we define a
 \emph{non-square} whitening matrix $\bR$ and the correlation vector $\bk_n$ as follows 
\begin{align}\label{eq:csquared}
\bR &\triangleq \begin{pmatrix}
\bA & \bB
\end{pmatrix} \,, \quad \bk_n \triangleq \bR^T \bK_{\bu, \bu}^{-1} \bk^*_{\bu, n}.
\end{align}
One can verify such $\bR$ and $\bk_n$ satisfy
Equation~\ref{eq:whitened-criteria}, thus offering a valid whitening strategy.
We note that since $\bR$ is non-square and $\bu = \bR \bepsilon$,
this strategy doubles the number of variational parameters
in each dimension of the whitened space.

Now we address how to efficiently compute $\bk_n$  defined in Equation~\ref{eq:csquared}. We first compute the intermediate quantity $\bk_n'= \bK_{\bu, \bu}^{-1} \bk^*_{\bu, n}$ via CG in $O(M \ln M)$ time. We then compute $\bk_n = \bR^\top \bk_n'$. Note that $\bR^{T}$ is embedded in the matrix $\bC^{1/2}$ which also admits the FFT diagonalization (Equation \ref{eq:csquared-fft}). Hence,  MVM with $\bR^\top$ can be also done in $O(M \ln M)$ time.  

Lastly, we show how to make CG's computation of $\bK_{\bu, \bu}^{-1} \bk^*_{\bu, n}$ faster with a well-structured preconditioner. 

\paragraph{Efficient Preconditioner}
\label{sec:toeplitz-preconditioner}
The ideal preconditioner $\bP$ is a matrix that whitens the matrix to be
inverted --- the ideal $\bP$ is $\bK_{\bu, \bu}^{-1}$.  However, we cannot
efficiently compute $\bK_{\bu, \bu}^{-1}$. But due to the convenient
diagonalization of the circulant embedding matrix $\bC$, we can
efficiently compute the inverse of $\bC$: 
\begin{align}
    \bC^{-1} = \bF^\top \bD^{-1} \bF \, .
\end{align}
Note that the upper left block of
$\bC^{-1}$ \emph{does not} correspond to $\bK_{\bu, \bu}^{-1}$
as we explicitly write out
\begin{align}
    \bC^{-1} =
    \begin{pmatrix}
    \left( \bK_{\bu, \bu} - \tilde{\bK} \bK_{\bu, \bu}^{-1} \tilde{\bK}^{\intercal}\right)^{-1} & ... \quad\\
    ... & ... \quad
    \end{pmatrix} \,\,.
\end{align}

However, when the number of inducing points are large enough,
$\bK_{\bu,\bu}$ approaches a banded matrix,
and so $\tilde{\bK}$ is increasingly sparse.
Therefore, the upper left block of $\bC^{-1}$ would be close to $\bK_{\bu, \bu}^{-1}$, suggesting that it can serve as an effective \emph{preconditioner} within PCG,
and therefore an effective strategy for solving a linear system with the 
kernel matrix. 
We note that this banded property is often exploited in developing
effective preconditioners \citep{chan1996conjugate, saad2003iterative}.
To justify this intuition, we anlayze the PCG convergence speed under various settings of 
kernel functions and inducing point densities in appendix.
We compare the performance of PCG and CG in systems of varying size
in Section~\ref{sec:experiments-preconditioner}.
We find that PCG
converges faster than CG across all systems, 
taking only a fraction of the number
of iterations that standard CG requires to converge.
This speedup is crucial --- PCG is a subroutine we use to compute the gradient
term corresponding to each observation $n$.

\paragraph{Summary of Fast Computation for $\bk_n$}
To summarize, we exploit additional computational benefits of
the hierarchical Toeplitz matrix through its circulant embedding matrix, which enables fast matrix square-root and matrix inverse. We further utilize these fast operations to design novel whitening and preconditioning strategies. 
Thus, the whitened correlation term $\bk_n = \bR^T \bK_{\bu, \bu}^{-1} \bk^*_{\bu, n}$
can be efficiently processed as follows:
\begin{enumerate}
  \item embed $\bK_{\bu, \bu}$ into a larger circulant matrix $\bC$;
  \item solve $\bK_{\bu, \bu} \bk'_n = \bk^*_{\bu, n}$ for the  intermediate term $\bk'_n$ with PCG,
  where we utilize the FFT diagonalization of $\bC$ and $\bC^{-1}$;
  \item compute $\bk_n = \bR^\top \bk'_n$ , where we utilize the FFT diagonalization of $\bC^{1/2}$.
\end{enumerate}
The space and time complexity of this procedure are $O(M)$ and $O(M \ln M)$.
This offers a speed-up over the Cholesky decomposition which has $O(M^2)$ space and $O(M^3)$ time complexity, respectively.
In Section~\ref{sec:experiments-whitened}, we examine this acceleration by comparing the time of computing $\bk_n$ using Cholesky and using HIP-GP, as the system size $M$ varying from $10^3$ to $10^6$. HIP-GP's strategy outperforms Cholesky for small values of $M$, and scales to larger $M$ where Cholesky is no longer feasible. We present HIP-GP's algorithmic details in appendix.

%% file: sections/method-variational-approx.tex
\subsection{Structured Variational Approximation}
\label{sec:variational-families}
Finally, we turn to the second bottleneck: how to represent
and manipulate variational parameters of mean $\bm$ and covariance $\bS$.
We propose the \emph{block independent} variational family
\begin{align}
  q(\bu) &= \prod_{b}^B \mathcal{N}(\bu_{b} \given \bm_{b}, \bS_{b}) \,,
  \label{eq:block-cov}
\end{align}
where $\bu_{b}$ denotes a subset of inducing points of size $M_b < M$
and $\bS_b$ is the $M_b \times M_b$ variational covariance for that
subset. Note that when $M_b=1$, it reduces to the $\emph{mean-field}$ variational family,
and when $M_b = M$, it is the \emph{full-rank} variational family.
Calculations of the inverse and log-determinant of block independent $\bS$ scale $O(B M_b^3)$
--- we must choose $M_b$ to be small enough to be practical.

We note that independence in the posterior is a more reasonable approximation
constraint \emph{in the whitened parameterization} than the original space.
The original GP prior, $p(\bu)$, is designed to have high correlation, and therefore
data are unlikely to decorrelate inducing point values.
In the whitened space, on the other hand, the prior is already uncorrelated.
Hence the whitened posterior
is not spatially correlated as much as the original posterior.
This is in addition to the benefits of optimizing in the whitened space
due to better conditioning.

\paragraph{Constructing Blocks}
The block independent approximation of Equation~\ref{eq:block-cov}
requires assigning inducing points to $B$
blocks.  Intuitively, blocks should include nearby points, and so we focus on
blocks of points that tile the space.
To reconcile the Toeplitz ordering and the block orderings (they may not be the same),
we simply have to permute any $M$-length vector
(e.g.~$\bk^*_{\bu,n}$ or $\bm$) before multiplication with $\bS$ and then undo
the permutation after multiplication.  Fortunately, this permutation is linear
in $M$.

%% file: sections/related-work.tex
\section{RELATED WORK}
\paragraph{Inter-domain GPs} The idea of the inter-domain Gaussian processes has been discussed
in \citep{lazaro2009inter, van2020framework}.
However, their primary interests are 
using inter-domain transformations to define inducing variables for specifying GP approximations, whereas our work explores the usage of SVGP framework to perform scalable modeling and inference
with inter-domain observations.

\paragraph{Scalable Inducing Point Methods}
We note several recent approaches to scaling
the number of inducing points in GP approximations.
\citet{shi2020sparse} takes an orthogonal strategy to ours by approximating GP with inducing points in two independent directions, whereas HIP-GP requires inducing points to densely cover the input space. However, while improved over standard SVGP, their method still remains a cubic complexity.
\citet{izmailov2018scalable} introduces
the tensor train decomposition into the variational approximation.
Alternatively, \citet{evans2018scalable} directly approximate the kernel with a
finite number of eigenfunctions evaluated on a dense grid of inducing points.
Both methods rely on \emph{separable} covariance kernels
to utilize the Kronecker product structure. This limits the
class of usable kernels.
The Matérn kernel, for example, is not separable across dimensions.
To fill that gap, we instead focus on the class of \emph{stationary kernels}.  

Another line of inducing point work is based on \emph{sparse kernel interpolations}.
KISS-GP uses a local kernel interpolation of inducing points
to reduce both the space and time complexity to $O(N+M^2)$ \citep{wilson2015kernel}.
SV-DKL also uses local kernel interpolation, and
exploits separable covariance structures and deep learning techniques
to address the problem of multi-output classification \citep{wilson2016stochastic}.
But these kernel interpolation methods are not applicable to
inter-domain observations under transformations.
More specifically, for standard (non-inter-domain) problems,  kernel interpolation methods approximate the $N \times N$ covariance matrix $\mathbf{K}_{\mathbf{N,N}}$ with $\mathbf{W} \mathbf{K}_{\mathbf{u, u}}\mathbf{W}^T$, where $\mathbf{W}$ is a sparse interpolation weight matrix. 
 However, for problems with integral observations, we must compute the integrated kernel $\mathbf{K}^{**}_{\mathbf{N,N}} = [\int \int Cov(\rho(\mathbf{x}_i), \rho(\mathbf{x}_j)) d\mathbf{x}_i d \mathbf{x}_j]_{i,j=1}^N$.
  Approximating this integral with local interpolation is not straightforward, and computing every integrated cross-covariance term is costly.
  Alternatively, HIP-GP decouples observations and inducing points into different domains
  through the \textit{inter-domain prior} (Equation~\ref{eq:inter-domain-GP}).
  This decoupled prior enables mini-batch processing of $k_{n,n}^{**}$, eliminates the need to compute cross-covariance terms  $k_{{n_i}, {n_j}}^{**}$,
  while still maintaining structure exploitation of $\mathbf{K}_{\mathbf{u,u}}$. 
  
 \paragraph{Fast Whitening Strategy} As is mentioned before, the classical whitening strategy is the Cholesky decomposition that has $O(M^2)$ space and $O(M^3)$ time complexity. 
 \citet{pleiss2020fast} provides a more general purpose method for fast matrix roots and is in particular applicable to whitening GP. Their method is an MVM-based  approach that leverages the contour integral quadrature and  requires $O(M \log M+QM)$ time for $Q$ quadrature points. Our whitening strategy specifically targets gridded inducing points and achieves more complexity savings ($O(M\log M)$ time). 
  

%% file: sections/experiments.tex
\section{EXPERIMENTS}
Due to the mismatch in missions of different scalable GP methods,
we focus most of our empirical study on HIP-GP and SVGP (with Cholesky whitening)
which serve the most similar purposes. We also include in appendix a standard GP problem on a UCI benchmark dataset \citep{Dua:2019} where we compare HIP-GP to exact GP \citep{wang2019exact}, sparse Gaussian Process regression (SGPR) \citep{titsias2009variational} and SVGP.

\input{sections/experiment-preconditioner.tex}

\input{sections/experiment-whitened.tex}

\input{sections/experiment-derivative-gp.tex}

\input{sections/experiment-housing.tex}

\input{sections/experiment-domain.tex}

%% file: sections/experiment-preconditioner.tex
\subsection{Effect of the Preconditioner}
\label{sec:experiments-preconditioner}
We first examine the effect of the preconditioner developed in Section
~\ref{sec:toeplitz-preconditioner}.
We run CG and PCG with the preconditioner for systems
of size $M=625~(25\times25)$, $M=2{,}500~ (50\times50)$, and $M=10{,}000~(100
\times 100)$ determined by a two-dimensional grid applied to the Mat\'ern
kernel.  We run the algorithm to convergence (at tolerance 1e-10) for $25$
randomly initialized vectors of size $M$.  We record the error at each
iteration --- the norm of the distance between the current solution and the
converged solution.

\begin{figure}[t!]
  \centering
  \includegraphics[width=.8\columnwidth]{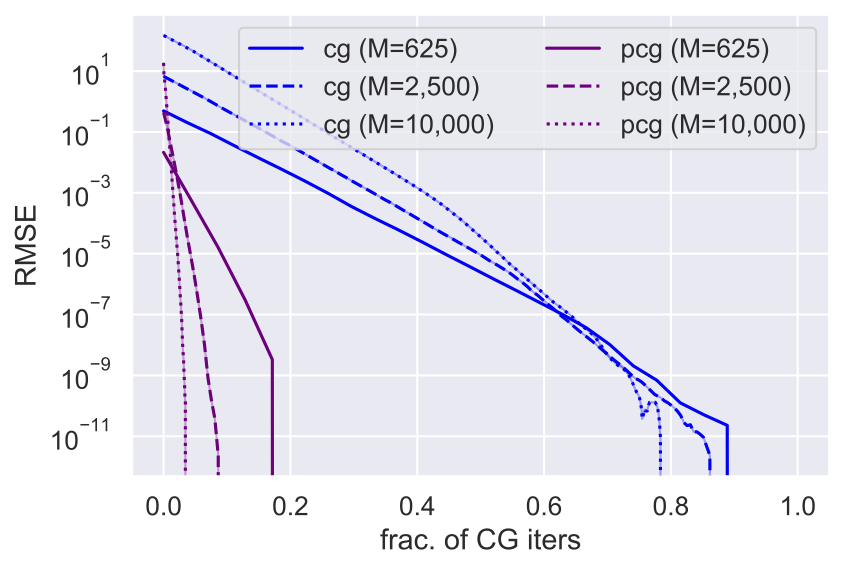}
  \caption{Convergence result of PCG v.s. CG.
	  We compare PCG to standard CG for systems of size $M=625, 2{,}500$
    and $10{},000$ over 25 independent runs.  We report RMSE as a
	  function of the fraction of \emph{total CG iterations} (to converge).
    PCG converges faster than CG, and for larger $M$ fewer
	  iterations are required.
	}
  \label{fig:preconditioner}
\end{figure}

We report the RMSE at each iteration in Figure~\ref{fig:preconditioner}.
We rescale the $x$-axis to run from 0 to 1 for each system of size $M$.
From this experiment we see two results: the Toeplitz
preconditioner is extremely effective and the preconditioner seems to be more
effective as the system size \emph{becomes larger}.
The fraction of CG iterations required for PCG to converge for
$M=10{,}000$ ($<4.5\%$) is much smaller than the fraction of iterations
required for $M=625$ ($<18\%$) to converge.
Without this preconditioner, we would expect each HIP-GP iteration
to take over twenty times longer to achieve similar precision.

%% file: sections/experiment-whitened.tex
\subsection{Speedup over Cholesky Decomposition}
\label{sec:experiments-whitened}
We examine the speedup of HIP-GP's whitening strategy over the
Cholesky whitening strategy in standard SVGP,
by comparing the time for solving the correlation term $\bk_n$.
We generate 200 random 1D observations, and
evenly-spaced inducing grids of size $M$ ranging from $10^3$ to $10^6$.
We apply a set of kernels including the Matérn kernels with $\nu = 0.5, 1.5, 2.5$ and the
squared exponential kernel. The marginal variance is fixed to $0.1$ for all $M$.
The length scale is set to $L/M$ where $L$ is the range of the data domain to utilize the inducing points efficiently.
The PCG within the HIP-GP algorithm is run to convergence at tolerence 1e-10.
The Cholesky decomposition is only available up to $M=10^4$ due to the memory limit.
All experiments are run on a NVIDIA Tesla V100 GPU with 32GB memory.

We report the wall clock time of computations applied to Matérn ($2.5$) kernel in Tabel~\ref{tab:whitened-time}.
The full report for all settings is presented in appendix.
HIP-GP's whitening strategy is consistently  faster than the 
 Cholesky whitening strategy across all experiments, and scales to larger $M$.

\begin{table}[t!]
  \centering
  \scalebox{.8}{\input{figures/whitening/Mat52-wallclock.tex}}
  \caption{Whitening time comparison (second) of HIP-GP v.s. SVGP with Matérn($2.5$) kernel.}
  \label{tab:whitened-time}
\end{table}

%% file: figures/whitening/Mat52-wallclock.tex
\begin{tabular}{lrrrrr}
\toprule
$M$ &     $10^3$       & $10^4$     & $10^5$     & $10^6$  \\
\midrule
HIP-GP   &  $\mathbf{0.0045}$  &  $\mathbf{0.0185}$  &  $\mathbf{0.3475}$  &  $\mathbf{1.4595}$  \\
SVGP     &  $0.0175$ &  $0.1745$  &   n/a      & n/a       \\
\bottomrule
\end{tabular}

%% file: sections/experiment-derivative-gp.tex
\subsection{Synthetic Derivative Observations}
\label{sec:derivative}
\begin{figure}[t!]
\centering
\begin{subfigure}{\columnwidth}
  \includegraphics[width=\textwidth]{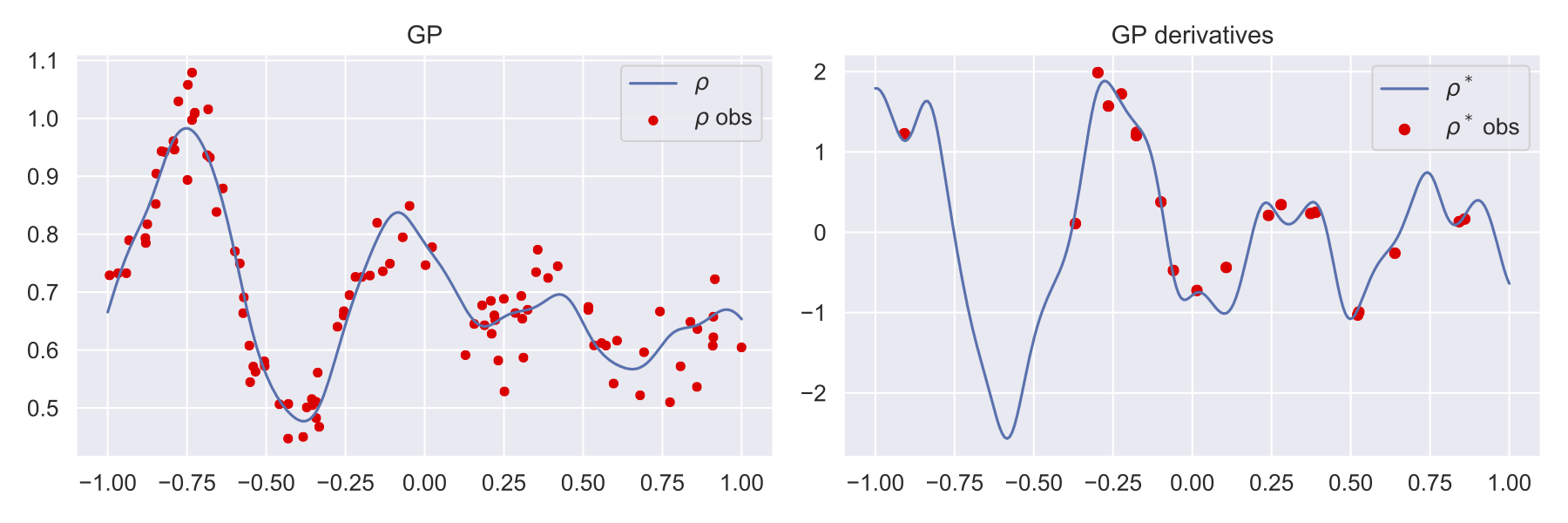}
  \caption{Synthetic function and derivative observations}
	\label{fig:derivative-observations}
  \vspace{0.25cm}
\end{subfigure}
\begin{subfigure}{\columnwidth}
  \includegraphics[width=\textwidth]{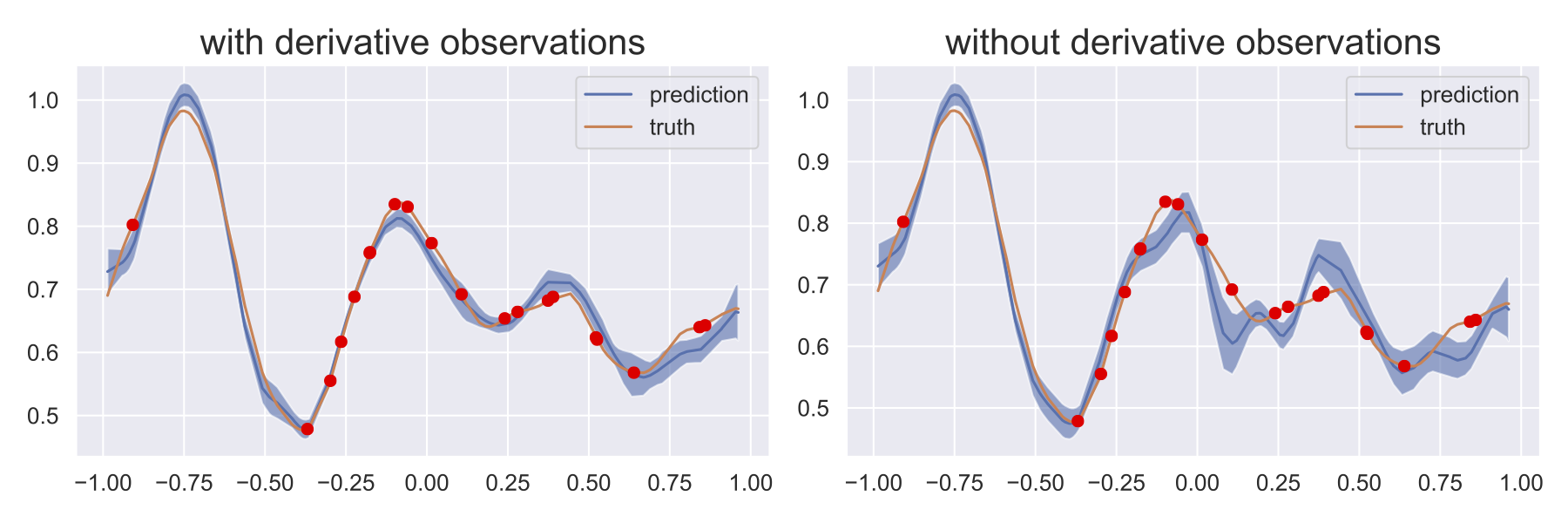}
  \caption{HIP-GP posterior prediction with / without derivative observations for 100 test data.
  The solid blue line is the mean prediction
  and the shaded blue area is the 1 posterior standard deviation band.
  The red points indicate derivative observation locations with true function values.}
	\label{fig:comparison-derivative}
  \vspace{0.25cm}
\end{subfigure}
\begin{subfigure}{\columnwidth}
  \centering
  \scalebox{.8}{
  \input{figures/derivative-synthetic/result.tex}
  }
	\caption{Predictive RMSE and uncertainty (i.e. average standard deviation) for 100 test data.}
	\label{tab:derivative-result}
\end{subfigure}
\caption{GP with derivative observations analysis.}
\label{fig:derivative-gp}
\end{figure}

To validate our inter-domain SVGP framework, we study a derivative GP problem.
We follow the work in \citet{solak2003derivative}, which introduces derivative observations
in addition to regular function observations to reduce
uncertainty in learning dynamic systems.
We synthesize a 1D GP function from a random neural network with
sinusoidal non-linearities, and
obtain function derivatives using automatic differentiation.
The total observations consist of 100 function observations
and 20 derivative observations,
with added noise level $=0.05$ and $0.2$ respectively,
as depicted in Figure~\ref{fig:derivative-observations}.

We compare two inter-domain SVGP framework-based methods, HIP-GP and the standard SVGP,
to the exact GP. We use the squared exponential kernel
with signal variance $0.5$ and length scale $0.1$. For both HIP-GP and SVGP,
we apply the full-rank variational family.
The maximum number of PCG iterations within HIP-GP is set to 20.
We evaluate the predictive performance on 100 test data.
From Figure~\ref{fig:comparison-derivative} and \ref{tab:derivative-result},
we see that the inter-domain SVGP framework successfully utilizes
the derivative observations to improve the prediction quality with reduced uncertainty,
and is comparable to the exact method.

%% file: figures/derivative-synthetic/result.tex
\begin{tabular}{llll}
\toprule
            & HIP-GP & SVGP   & Exact GP \\
\midrule
RMSE        & 0.0192 & 0.0192 & 0.0192 \\
Uncertainty & 0.0198 & 0.0206 & 0.0198   \\
\bottomrule
\end{tabular}

%% file: sections/experiment-housing.tex
\subsection{Spatial Analysis: UK Housing Prices}
Now we test HIP-GP on a standard GP problem,
i.e., the transformation $\mathcal{L}$ is an identity map.
We apply HIP-GP to (log) prices of apartments as a function of latitude and
longitude in England and Wales\footnote{HM land registry price paid data
available
\href{https://ckan.publishing.service.gov.uk/dataset/land-registry-monthly-price-paid-data}
{here}.}.  The data include 180{,}947 prices from 2018, and we train on
160{,}947 observations and hold out 20{,}000 to report test error.
We use the standard SVGP as baseline.

\begin{figure}
\centering
\begin{subfigure}{.48\columnwidth}
  \includegraphics[width=\textwidth]{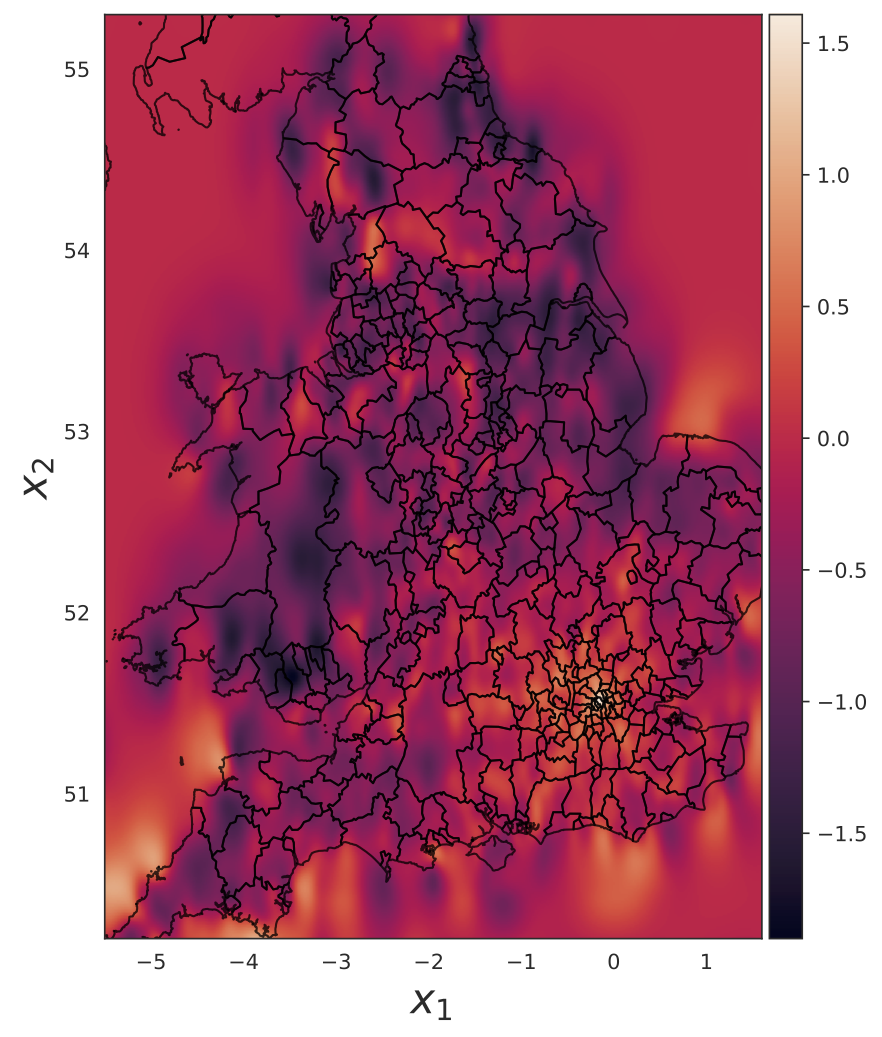}
  \caption{Posterior mean}
	\label{fig:housing-mu}
\end{subfigure}
~
\begin{subfigure}{.48\columnwidth}
  \includegraphics[width=\textwidth]{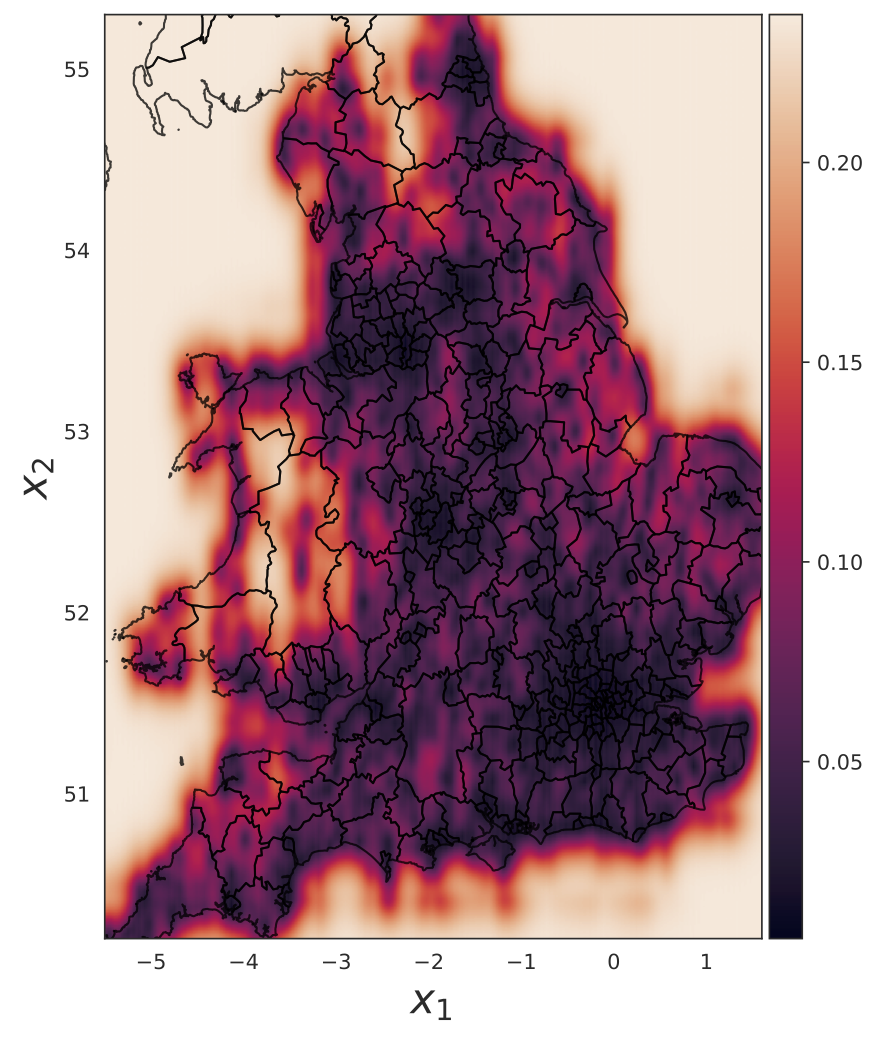}
  \caption{Posterior st.~dev.}
  \label{fig:housing-sig}
\end{subfigure}


\begin{subfigure}{\columnwidth}
  \centering
  \scalebox{.7}{
  \input{figures/uk-housing/uk-more-inducing-point.tex}
  }
	\caption{Top row: predictive RMSE. Bottom row: average training time (second) per epoch.}
	\label{tab:housing-table}
\end{subfigure}
\caption{UK Housing Analysis}
\label{fig:uk-housing}
\end{figure}

\paragraph{Scaling Inducing Points}
We run HIP-GP on an increasingly dense grid of inducing points $M$.
In all experiments, we use the Matérn ($2.5$) kernel
 and apply the block-independent variational family with
neighboring block size $M_b = 100~(10\times10)$  for HIP-GP and SVGP.
The maximum number of PCG iterations within HIP-GP is set to 20 and 50
for training and evaluation.
The predictive performance measured by RMSE and the training time are 
displayed in Figure~\ref{tab:housing-table}.
From this result, we conclude that
(i) increasing $M$ improves prediction quality;
(ii) the performance of HIP-GP is almost indistinguishable to that of
SVGP given the same $M$. (iii) Again, HIP-GP runs faster than SVGP and scales to larger $M$. 
The best prediction of HIP-GP is
depicted in Figure~\ref{fig:housing-mu} and \ref{fig:housing-sig}.

\begin{figure}[t!]
  \centering
  \includegraphics[width=.9\columnwidth]{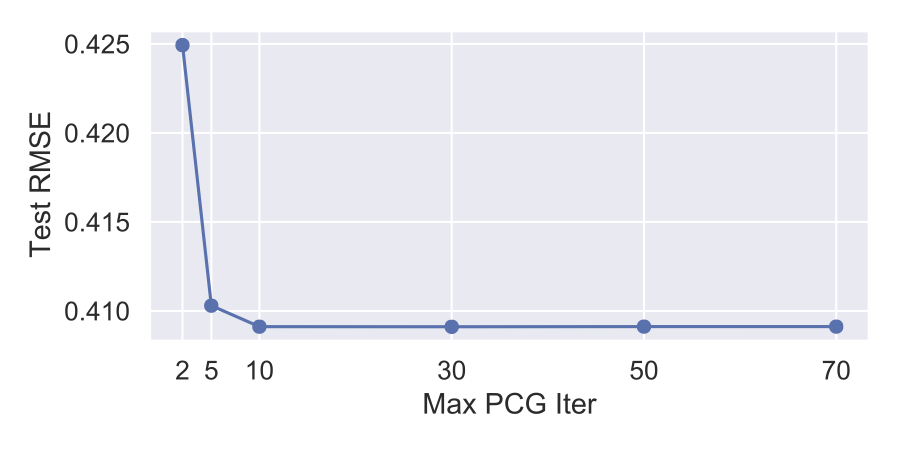}
  \caption{Stochastic optimization is robust to early stopping
     of PCG iterations.
  }
  \label{fig:pcg-iter}
\end{figure}

\paragraph{PCG Iteration Early Stopping}
Additionally, we examine the effect of the maximum number of PCG iterations
when computing $\bk_n$ on approximation
quality.  Figure~\ref{fig:pcg-iter} depicts test
RMSE as a function of PCG iteration for $M = 14{,}400$
on the test dataset of size $N=20{,}000$.
The final approximation quality is robust to the number of PCG iterations used.
The upshot is that HIP-GP needs only a small number of PCG iterations
to be effective. 

%% file: figures/uk-housing/uk-more-inducing-point.tex
\begin{tabular}{lrrrrrrr}
    \toprule
    M & 10{,}000 & 14{,}400 & 19{,}600 &  25{,}600 & 32{,}400 & 40{,}000 \\
    \midrule
    HIP-GP (RMSE)& 0.411 & 0.409 & 0.400 &0.397 &0.393 & 0.389 \\
    SVGP (RMSE) & 0.412 & 0.409 & 0.398 &0.396 & n/a & n/a \\
    \midrule
    HIP-GP (time)     &\textbf{91.2}  &\textbf{115.7} &\textbf{119.8} &\textbf{130.7} &\textbf{129.5} &\textbf{133.2} \\
    SVGP (time)   &193.6          &406.3 &668.1 &898.2 &n/a &n/a\\
    \bottomrule 
    \end{tabular}

%% file: sections/experiment-domain.tex
\subsection{Inferring Interstellar Dust Map}

\label{sec:experiment-dustmap}
\begin{figure}[t!]
  \centering
  \begin{subfigure}{.49\columnwidth}
    \includegraphics[width=\columnwidth]{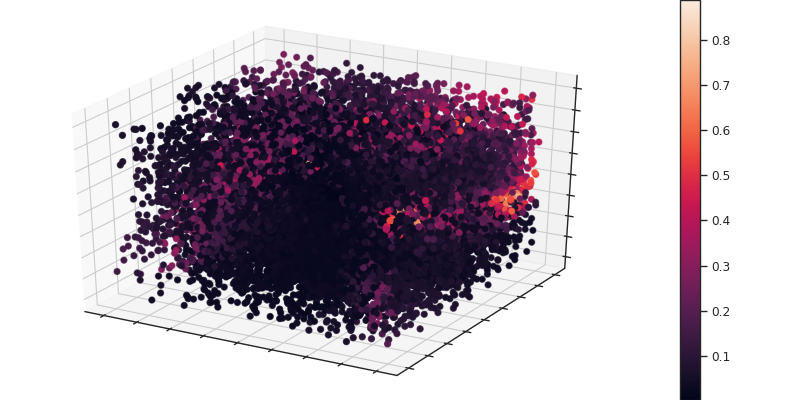}
    \caption{Posterior mean of $\rho^*$}
    \label{fig:domain-predict-rhostar}
    \vspace{.5cm}
  \end{subfigure}
  \begin{subfigure}{.49\columnwidth}
    \includegraphics[width=\columnwidth]{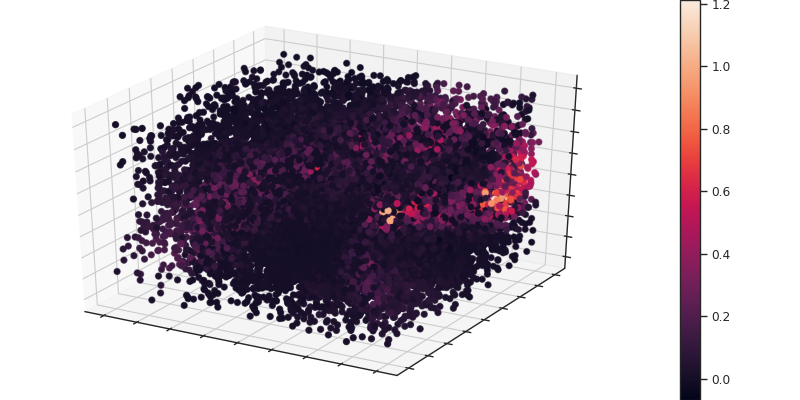}
    \caption{Posterior mean of $\rho$}
		\label{fig:domain-predict-rho}
    	\vspace{.5cm}
  \end{subfigure}
  \begin{subfigure}{\columnwidth}
		\centering
    \scalebox{.8}{
    \input{figures/domain/domain-report.tex}

    }
    \caption{Predictive statistics for integrated observations}
    \label{tab:domain-report}
  \end{subfigure}
	\caption{HIP-GP performance on 20{,}000 held-out data.
    Top: Posterior mean predictions in the integrated domain ($\rho^*$)
    and the latent domain ($\rho$).
    Bottom: We report the mean absolute error (MAE), the mean square error (MSE)
    and the test log likelihood in the intergrated domain.
  }
  \label{fig:domain-experiment}
\end{figure}

Finally, we investigate an inter-domain GP problem with $\mathcal{L}$ being the integral transformation:
inferring the interstellar dust map from integral observations.
The interstellar dust map $\rho$ is a three-dimensional density function at each location in the Galaxy.
The observations $y$, also known as the starlight extenctions, are noisy line
integrals of the dust function \citep{rezaei2017inferring}.
We experiment with the Ananke dataset,
which is comprised of $500{,}000$ starlight extinctions
within $4\textrm{kpc} \times 4 \textrm{kpc} \times 2 \textrm{kpc}$ region
 of a high resolution Milky Way like galaxy simulation --- a cutting edge simulation
 in the field because of the gas and dust resolution
  \citep{wetzel2016reconciling, hopkins2018fire, sanderson2020synthetic}.
Our goal is to infer the underlying dust map $\rho$
from the noisy extinctions $y$.

We compare HIP-GP with $M=62{,}500 \,(50\times 50 \times 25)$
and SVGP with $M=16{,}384 \,(32 \times 32 \times 16)$ -- the largest $M$ feasible.
For both methods, we apply the block-independent variational
parameterization with neighboring block size $M_b = 8~(2\times2\times2)$,
and the Matérn$\,(1.5)$ kernel.
The maximum number of PCG iterations within HIP-GP is set to
200 and 500 for training and evaluation.
We use Monte Carlo estimation to compute the
inter-domain and transformed-domain covariance functions
in Equations~\ref{eq:domain-kernel} and \ref{eq:domain-kernel2}.
We hold out 20{,}000 points for evaluation.
The posterior mean predictions of the extinctions $\rho^*$
and the latent dust map $\rho$ are displayed in
Figure~\ref{fig:domain-predict-rhostar} and \ref{fig:domain-predict-rho}.
The predictive test statistics are summarized in Table~\ref{tab:domain-report}.
We see that with more inducing points, the predictive accuracy is enhanced.
HIP-GP can scale to larger $M$ which enables better prediction quality,
while SVGP is limited to $M$ around $16{,}000$.

%% file: figures/domain/domain-report.tex
\begin{tabular}{lrrrrrr}
\toprule
{}                  &   MAE   &  MSE    & loglike  \\
\midrule
HIP-GP ($M=62{,}500$) & \textbf{0.0101} & \textbf{0.0012} & \textbf{2.3517} \\
SVGP ($M=16{,}384$)   & 0.0153 & 0.0020 &  2.0690 \\
\bottomrule
\end{tabular}

%% file: sections/discussion.tex
\section{DISCUSSION}
We formulate a general SVGP framework for inter-domain GP problems.
Upon this framework, we further scale the standard SVGP inference
by developing the HIP-GP algorithm,
with three technical innovations (i) a fast whitened parameterization,
(ii) a novel preconditioner for fast linear system solves with
hierarchical Toeplitz structure, and (iii)
a structured variational approximation.
The core idea of HIP-GP lies in the structured exploitations of the kernel matrix and the variational posterior.
Therefore, it can be potentially extended to various settings, e.g.
the case where a GP is a part
of a bigger probabilistic model,
and the non-Gaussian likelihoods thanks to recent advance
in non-conjugate GP inference \citep{salimbeni2018natural}.

Future works involve more in-depth analysis of
such CG-based approximate GP methods.
On the applied side, we will to apply HIP-GP to
the Gaia dataset \citep{gaia2018gaia}
which consists of nearly 2 billion stellar observations.

%% file: sections/acknowledgement.tex
\section*{Acknowledgements}
We thank the reviewers for their detailed feedback and suggestions. 
The interstellar dust map experiments in Section~\ref{sec:experiment-dustmap} were run on the Iron cluster at the Flatiron Institute, and we are grateful to the scientific computing team for their continual and dedicated technical assistance and support. The Flatiron Institute is supported by the Simons Foundation.

%% file: appendix-arxiv.tex
\onecolumn
\aistatstitle{Supplementary Materials:
Hierarchical Inducing Point Gaussian Processes for Inter-domain Observations}

\section{The HIP-GP Algorithm}

We describe two algorithms that are core to the  acceleration techniques we develop in Section 3.2. 
Algorithm~\ref{alg:toeplitz-mvm} computes a fast MVM
with a hierarchical Toeplitz matrix using the circulant embedding
described in Algorithm~\ref{alg:circulant-embedding}. 
Note that we can similarly compute the MVM $\bR^\top \bv$ simply by adapting Algorithm~\ref{alg:toeplitz-mvm} to perform FFT on $\bC^{1/2}$ instead of on $\bC$. Together these algorithms are sufficient for use within PCG to efficiently compute $\bk_n$.

\begin{algorithm}[!h]
  \KwData{
    $T$ ($N_1 \times ... \times N_D$ representation of hierarchical Toeplitz matrix);
  }
  \KwResult{$C$ (circulant embedding of $T$)}
  $C \leftarrow T$ \tcp*{copy}
  \For{$d\gets1$ \KwTo $D$}{
    $C_r \leftarrow \texttt{reverse-dim}(C, \text{dim}=d)$ \\ 
    $C_r \leftarrow \texttt{chop-single-dim}(C_r, \text{dim}=d)$ \\ 
    $C_r \leftarrow \texttt{binary-zero-pad}(C_r)$ \tcp*{front pad}
    $C \leftarrow \texttt{concat}(C, C_r, \text{dim}=d)$ \\
  }
  return $C$
\caption{Hierarchical circulant embedding.}
\label{alg:circulant-embedding}
\end{algorithm}

\begin{algorithm}[!h]
  \KwData{
    $\bk_0$ (first row of $\bK$ in $C$-order);
    $\bv$ (vector, also in $C$-order);
    $N_1, \dots, N_D$ (grid dimensions)
  }
  \KwResult{$\bK \bv$ (matrix-vector product)}
  $T \leftarrow \texttt{reshape}(\bk_0, N_{1:D})$ \tcp*{to $N_1 \times \dots \times N_D$}
  $V \leftarrow \texttt{reshape}(\bv, N_{1:D})$ \tcp*{to $N_1 \times \dots \times N_D$}
  $C \leftarrow \texttt{Circ-Embed}(T, N_{1:D})$ \tcp*{} 
  $V \leftarrow \texttt{Zero-Embed}(V, N_{1:D})$ \tcp*{match $C$}
  $res \leftarrow \texttt{ifft}( \texttt{fft}(C) \cdot \texttt{fft}(V) )$ \tcp*{$D$-dim \texttt{fft}}
  return \texttt{flatten}(res) \tcp*{flatten in \texttt{C}-order}
\caption{Matrix-vector multiplication $\bK \bv$ for a symmetric
  hierarchical Toeplitz matrix $\bK$ and vector $\bv$.
}
\label{alg:toeplitz-mvm}
\end{algorithm}

\section{Optimization Details}
In this section, we derive the gradients for structured variational parameters and kernel hyperparameters.
\subsection{Variational parameters}
The structured variational posterior is characterized by $N(\bm, \bS)=\prod_{i=1}^B \mathcal{N}(\bm_i, \bS_i)$,
where we decompose the $M\times M$ matrix $\bS$  into $B$ block-independent covariance matrices of block size $M_b$:
\begin{align}
  \bS = \begin{pmatrix}
& \bS_1 \\
& & \bS_2 \\
&&&\cdots \\
&&&& \bS_B
\end{pmatrix},
\end{align}
and the vector $\bm$ into corresponding $B$ blocks: $\bm_1, \bm_2, \cdots, \bm_B$.

\subsubsection{Direct solves}
We first consider directly solving the optimal $\bm$ and $\bS$.

Taking the derivatives of the HIP-GP objective w.r.t. $\bm$ and $\bS$, we obtain
\begin{align}
  \frac{\partial \mathcal{L}}{\partial \bS_i}
  &= -\frac{1}{2} \underbrace{\left((\sum_n \frac{1}{\sigma_n^2} \bk_{n,i} \bk_{n,i}^\top) + \bI_{M_b} \right)}_{\triangleq \bLambda_i} + \frac{1}{2} \bS_i^{-1}
  \,, \qquad \textrm{for } i = 1:B \\
\frac{\partial \mathcal{L}}{ \partial \bm} &= \underbrace{ \sum_n \frac{1}{\sigma_n^2} y_n \bk_n}_{\triangleq \bb}
    - \underbrace{\left(\sum_n \frac{1}{\sigma_n^2}\bk_n \bk_n^\top + \bI_M \right)}_{\triangleq \bLambda} \bm
\end{align}
where a vector or a matrix with subscript $i$, denotes its $i$-th block.

The optimum can be solved in closed form by setting the gradients equal to zero, i. e.
\begin{align}
  \frac{\partial \mathcal{L}}{\partial \bS_i} = 0 \qquad  \Rightarrow  \qquad &\bS_i = \bLambda_i^{-1}, \qquad \textrm{for } b=1:B \\
  \frac{\partial \mathcal{L}}{ \partial \bm} = 0 \qquad  \Rightarrow   \qquad &\bm   = \bLambda^{-1} \bb .
\end{align}

 If $M$ is very large, this direct solve will be infeasible.
But note that $\bLambda_i, \bb$ and $\bLambda$ are all summations over some data terms,
hence we can compute an unbiased gradient estimate using a small number of samples which is more efficient. We will use natural gradient descent (NGD) to perform optimization. 

\subsubsection{Natural gradient updates}
To derive the NGD updates, we need the other two paramterizations of $N(\bm, \bS)$, namely,
\begin{itemize}
  \item the \textit{canonical parameterization}: $ \{\btheta_{1, i}\}_{i=1}^B, \{\btheta_{2,i} \}_{i=1}^B$
  where $\btheta_{1,i}= \bS_i^{-1} \bm_i$ and $\btheta_{2,i} = -\frac{1}{2} \bS_i^{-1}$, $i=1:B$; and
  \item the \textit{expectation parameterization}: $\{\etab_{1,i} \}_{i=1}^B, \{\etab_{2,i} \}_{i=1}^B$
  where $\etab_{1,i} = \bm_i$ and $\etab_{2,i} = \bm_i \bm_i^\top + \bS_i$, $i=1:B$.
\end{itemize}

In Gaussian graphical models, the natural gradient for the \textit{canonical parameterization} corresponds to
the standard gradient for the \textit{expectation parameterization}. That is,
\begin{align}
  \frac{\partial}{\partial \etab} \mathcal{L} &= \tilde{\nabla}_{\btheta} \mathcal{L},
\end{align}
where $\tilde{\nabla}_\theta$ denotes the natural gradient w.r.t. $\theta$.

By the chain rule, we have

\begin{align}
  \frac{\partial \mathcal{L}}{ \partial \etab_{1,i}}
&= \frac{\partial \mathcal{L}}{\partial{\bm_i}} \frac{\partial \bm_i}{\partial \etab_{1,i}}
  + \frac{\partial \mathcal{L}}{\partial \bS_i}\frac{\partial \bS_i}{ \partial \etab_{1,i}} \\
&= \bb_i -\bS_i^{-1}\bm_i - [(\bLambda \bm )_i - \bLambda_i \bm_i ], \\
\frac{\partial \mathcal{L}}{\partial \etab_{2,i}}
&= \frac{\partial \mathcal{L}}{\partial{\bm_i}} \frac{\partial \bm_i}{\partial \etab_{2,i}}
   + \frac{\partial \mathcal{L}}{\partial \bS_i}\frac{\partial \bS_i}{ \partial \etab_{2,i}} \\
&= -\frac{1}{2}\bLambda_i + \frac{1}{2}\bS_i^{-1}.
\end{align}

Therefore, the natural gradient updates for $\btheta$ are as follows:
\begin{align}
  \btheta_{1,i} &\leftarrow \btheta_{1,i} + l \frac{\partial \mathcal{L}}{ \partial \etab_{1,i}}
   = \btheta_{1,i} + l \left( \bb_i -\bS_i^{-1}\bm_i - [(\bLambda \bm )_i - \bLambda_i \bm_i ] \right)   \\
  \btheta_{2,i} & \leftarrow \btheta_{2,i} + l \frac{\partial \mathcal{L}}{\partial \etab_{2,i}}
  = \btheta_{2,i} + l \left(  -\frac{1}{2}\bLambda_i + \frac{1}{2}\bS_i^{-1} \right),
\end{align}
where $l$ is a positive step size.

\subsection{Kernel hyperparameters}
We now consider learning the kernel hyperparameters $\btheta$ with gradient descent.

For the convenience of notation, we denote the gram matrix $\bK_{\bu, \bu}$ as $\bK$.
HIP-GP computes $\bK^{-1} \bv$ by PCG. 
Directly auto-differentiating through PCG may be numerically instable.
Therefore, we derive the analytical gradient for this part.
Denote $\bc$ as the first row of $\bK$ --- $\bc$ fully characterizes the symmetric Toeplitz matrix $\bK$.
It suffices to manually compute the derivative w.r.t. $\bc$, i.e.
 $\frac{\partial \br^\top \bK^{-1} \bv }{ \partial \bc}$,
since by the following term $\frac{\partial \bc}{\partial \btheta}$ can be taken care of with auto-differentiation.

We note the following equality
\begin{align}
 \frac{\partial \br^\top \bK^{-1} \bv }{ \partial \bc}
  &= -  (\bK^{-1} \br)^\top \frac{\partial \bK}{ \partial \bc} \bK^{-1} \bv.
\end{align}

The computation of $\bb = \bK^{-1}v$ is done in the forward pass and therefore can be cached for the backward pass.
Additional computations in the backward pass are (1) $\ba = \bK^{-1} \br$ and
 (2) $\frac{ \partial \ba^\top  \bK \bb}{ \partial \bc }$. (1) can be computed efficiently
 using the techniques developed in HIP-GP.
Now we present the procedure to compute (2):
\begin{align}
  \frac{\partial \ba^\top \bK \bb}{\partial \bc}  &= \sum_{ij} a_i b_j \frac{\partial K_{ij}}{\partial \bc} \\
  &= \sum_{ij} a_i b_j  \be_{|i-j|+1} \\
&= \textrm{toeplitz-mm} (b_1 \be_1, \bb, \ba) + \textrm{toeplitz-mm} (a_1\be_1, \ba, \bb) - (\ba^\top \bb) \be_1,
\end{align}
where $\be_{i}$ denotes the vector with a 1 in the $i$-th coordinate and 0's elsewhere,
and $\textrm{toeplitz-mm}(\bx,\by,\bz)$ denotes the Toeplitz MVM $\bT \bz$, with the
Toeplitz matrix $\bT$ characterized by its first column vector $\bx$ and first row vector $\by$
--- this Toeplitz MVM can be also efficiently computed via its circulant embedding.

\section{Additional Experiment Results}

\subsection{Empirical analysis on preconditioner}

\begin{figure}[t!]
\centering
\begin{subfigure}{0.8\columnwidth}
  \includegraphics[width=\textwidth]{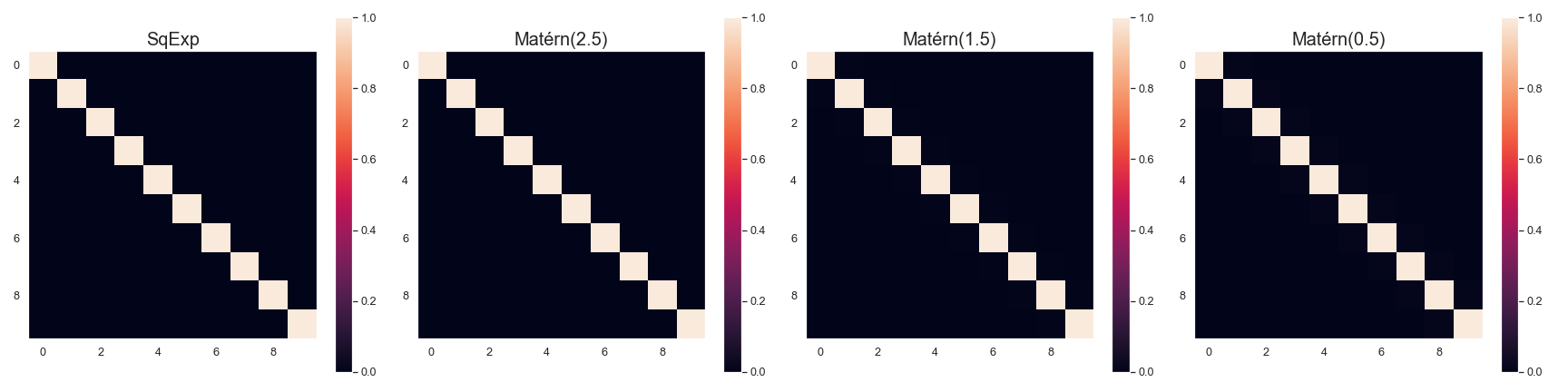}
  \caption{Inducing point kernel matrix $\bK_{\bu, \bu}$ with $M=10$}
	\label{fig:ell5e-2-M10}
  \vspace{0.25cm}
\end{subfigure}
\begin{subfigure}{0.8\columnwidth}
  \includegraphics[width=\textwidth]{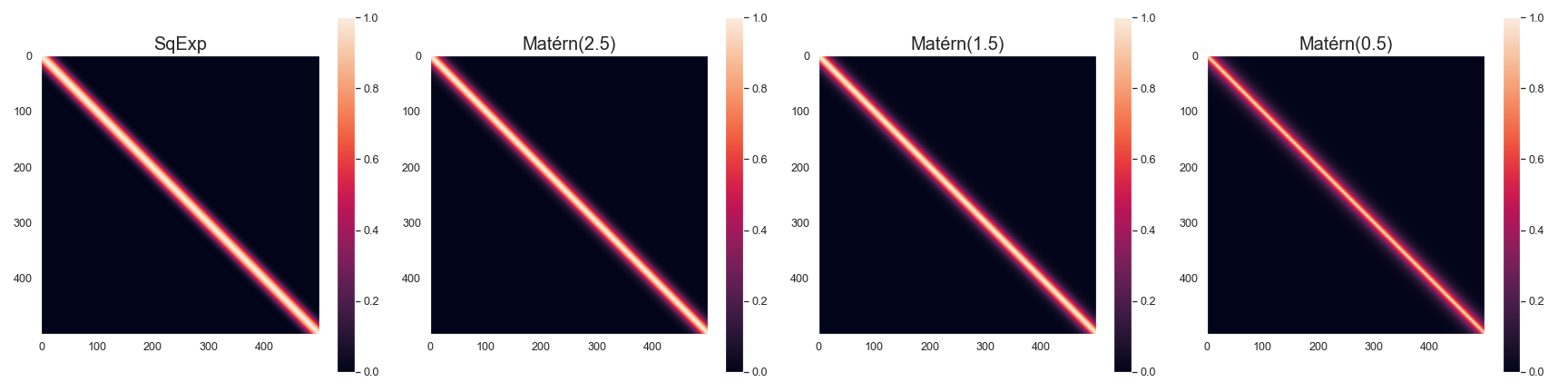}
  \caption{Inducing point kernel matrix $\bK_{\bu, \bu}$ with $M=500$}
	\label{fig:ell5e-2-M500}
  \vspace{0.25cm}
\end{subfigure}
\begin{subfigure}{0.8\columnwidth}
  \includegraphics[width=\textwidth]{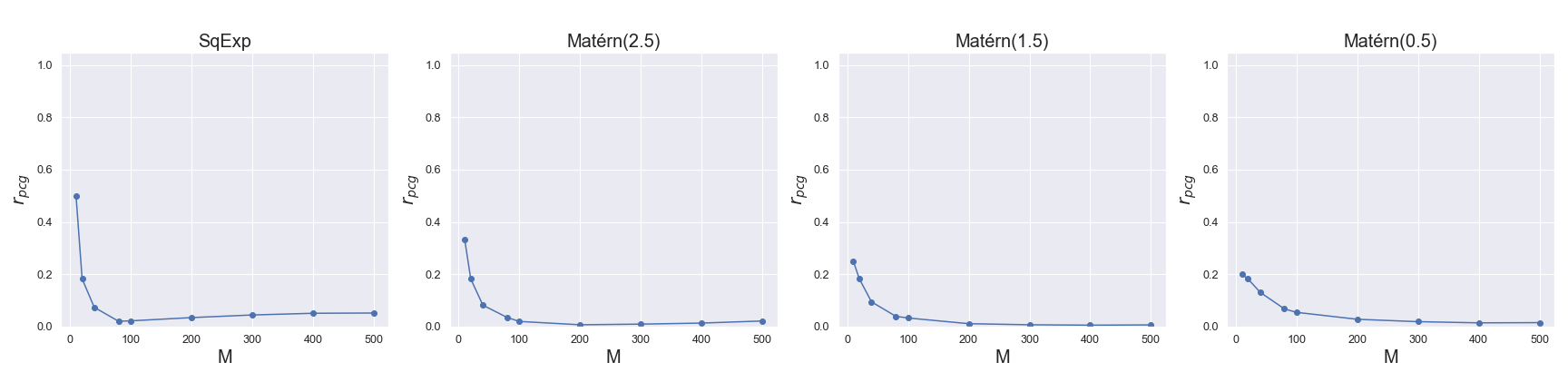}
  \caption{$r_{pcg}$ v.s. $M$ for different kernels. }
	\label{fig:ell5e-2-convergence}
  \vspace{0.25cm}
\end{subfigure}
\caption{Empirical analysis for preconditioner. The kernel lengthscale is 0.05. }
\label{fig:preconditioner-ell=5e-2}
\end{figure}

In this section, we present an empirical analysis on the preconditioner developed in Section~\ref{sec:toeplitz-preconditioner}.
Specifically, we investigate our intuition
on the ``banded property" that makes the preconditioner effective:
when the number of inducing points $M$ is large enough,
the inducing point Gram matrix $\bK_{\bu, \bu}$ is increasingly sparse,
and therefore the upper left block of $\bC^{-1}$ will be close to $\bK_{\bu, \bu}^{-1}$.

\begin{figure}[t!]
\centering
\begin{subfigure}{0.8\columnwidth}
  \includegraphics[width=\textwidth]{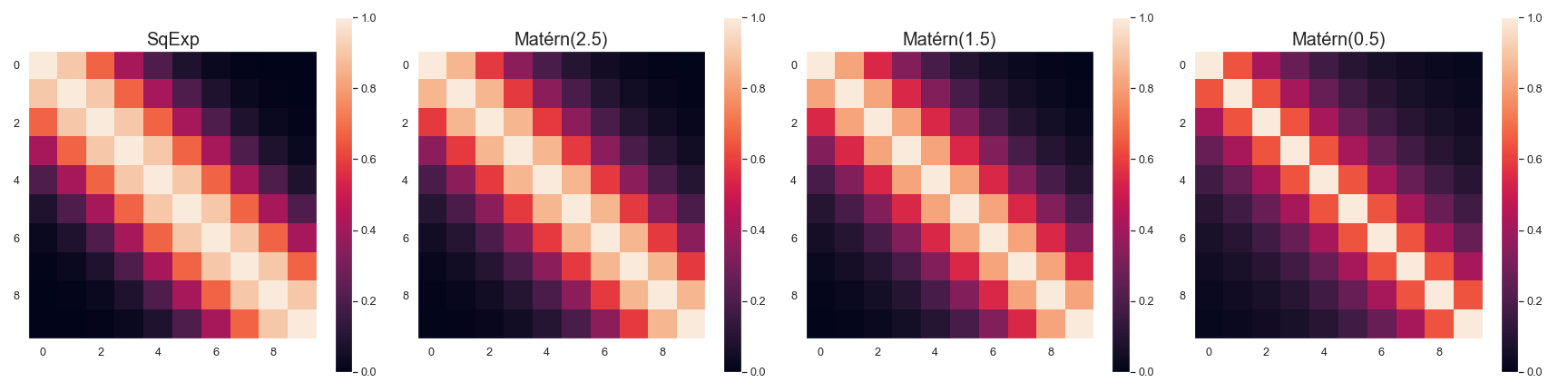}
  \caption{Inducing point kernel matrix $\bK_{\bu, \bu}$ with $M=10$}
	\label{fig:ell5e-1-M10}
  \vspace{0.25cm}
\end{subfigure}
\begin{subfigure}{0.8\columnwidth}
  \includegraphics[width=\textwidth]{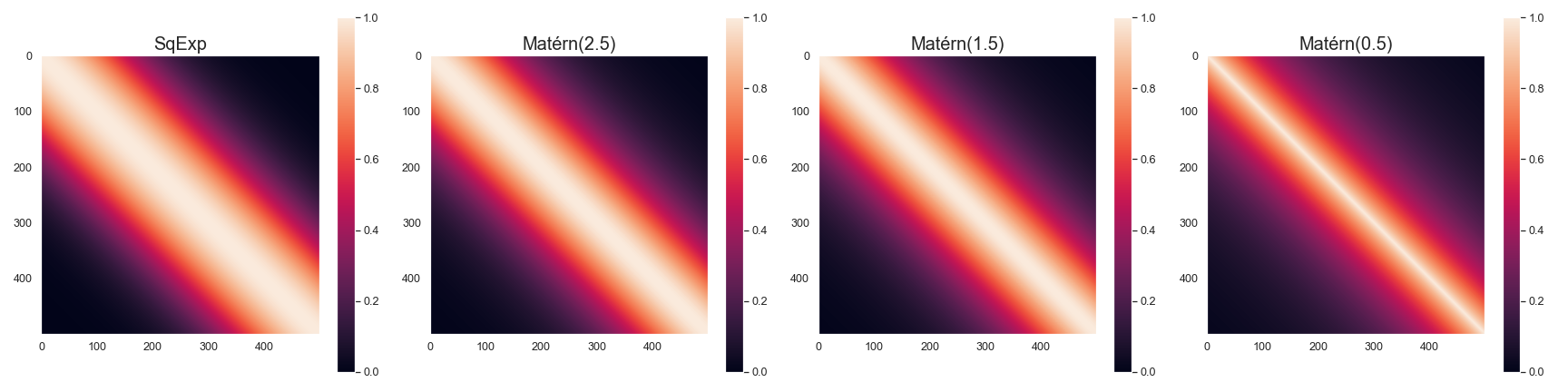}
  \caption{Inducing point kernel matrix $\bK_{\bu, \bu}$ with $M=500$}
	\label{fig:ell5e-1-M500}
  \vspace{0.25cm}
\end{subfigure}
\begin{subfigure}{0.8\columnwidth}
  \includegraphics[width=\textwidth]{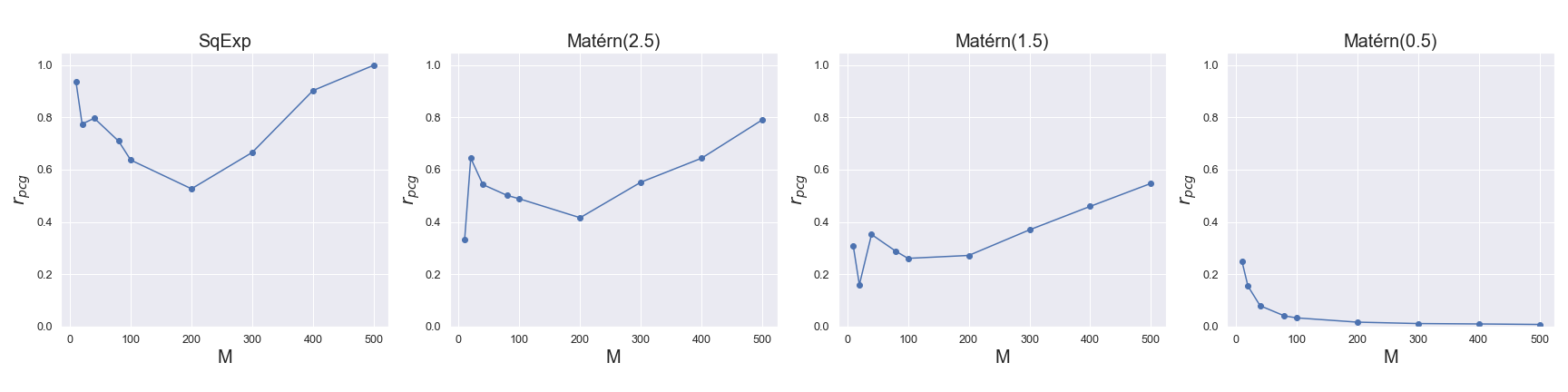}
  \caption{$r_{pcg}$ v.s. $M$ for different kernels. }
	\label{fig:ell5e-1-convergence}
  \vspace{0.25cm}
\end{subfigure}
\caption{Empirical analysis for preconditioner. The kernel lengthscale is 0.5. }
\label{fig:preconditioner-ell=5e-1}
\end{figure}

We note that the sparsity of the kernel matrix $\bK_{\bu, \bu}$ depends on three factors
(1) $M$, the number of inducing points, (2) $l$, the lengthscale of the kernel,
and (3) the property of the kernel function itself such as smoothness. 
To verify our intuition, we conduct the PCG convergence experiment
by varying the combinations of these three factors.
We evenly place $M$ inducing points in the $[0,2]$ interval that form the Gram matrix $\bK_{\bu, \bu}$,
and randomly generate 25 vectors $\bv$ of length $M$. We vary $M$ ranges from 10 to 500, and experiment with 4 types of kernel function: squared exponential kernel,
Matérn (2.5), Matérn (1.5) and Matérn (0.5) kernels.
For all kernels, we fix the signal variance $\sigma^2$ to  1
and the lengthscale $l$ to 0.05 and 0.5 in two separate settings.
We run CG and PCG to solve $\bK_{\bu, \bu}^{-1} \bv$
up to convergence with tolerance rate at 1e-10.
We compare the fraction of \# PCG iterations required for convergence over
\# CG iterations required for convergence, denoted as $r_{pcg}$.
The results are displayed in Figure~\ref{fig:preconditioner-ell=5e-2} (for $l=0.05$)
and Figure~\ref{fig:preconditioner-ell=5e-1} (for $l=0.5$).

Figure~\ref{fig:ell5e-2-M10} - \ref{fig:ell5e-2-M500}
and Figure~\ref{fig:ell5e-1-M10} - \ref{fig:ell5e-1-M500}
depict the kernel matrix $\bK_{\bu, \bu}$ for $M=10$ and $M=500$
with $l=0.05$ and $0.5$, respectively.
Figure~\ref{fig:ell5e-2-convergence} and \ref{fig:ell5e-1-convergence}
plot $r_{pcg}$ over $M$ for different kernels and lengthscales.
From these plots, we make the following observations:

\begin{enumerate}[(1)]

   \item $r_{pcg}$ are consistently smaller than 1,
   which verifies the effectiveness of the preconditioner.

   \item In most cases, PCG converges faster when the system size $M$ is bigger. For example, $r_{pcg}$ decreases as $M$ increases in Figure~\ref{fig:ell5e-2-convergence} where $l=0.05$.  However, we note that 
     PCG convergence can be slowed down when the system size $M$ exceeds certain threshold in some cases (e.g.first three plots of Figure~\ref{fig:ell5e-1-convergence} where $l=0.5$). 
    To see why this happens, we compare the plots of kernel matrices with $l=0.5$ 
    for $M=10$ and $M=500$ in Figure~\ref{fig:ell5e-1-M10} and \ref{fig:ell5e-1-M500}.
    Since the lengthscale $l=0.5$ is relatively large
    with respect to the input domain range, the resulting $\bK_{\bu, \bu}$ for $M=10$ is sparse enough to approach a diagonal matrix. 
    However, when we increase $M$ to 500,
    1 lengthscale unit covers too many inducing points, 
    making $\bK_{\bu, \bu}$ less ``banded" and the preconditioner less effective. 
    This observation is also consistent with our intuition.
    
      \item For kernels that are less smooth, the PCG convergence
   speed-ups over CG are bigger given the same $M$,
   e.g.  Matérn (0.5) kernel has smaller $r_{pcg}$
   than squared exponential kernel does under the same configuration.
   We also observe that $\bK_{\bu, \bu}$ with Matérn (0.5) kernel is more diagonal-like
   than $\bK_{\bu, \bu}$ with squared exponential kernel from the kernel matrix plots.
   Together with (2),
   these results show that when the kernel matrix is more banded, the PCG convergence is accelerated more.
\end{enumerate}

In conclusion, the PCG convergence speed depends
on the ``banded" property of the inducing point kernel matrix $\bK_{\bu, \bu}$,
which further depends on $M$ and the smoothness of the kernel. As $\bK_{\bu, \bu}$ approaches a banded matrix,
the preconditioner speeds up convergence drastically. 

\subsection{Additional experiment results for Section 5.2}
We include additional experiment results on the other 3 kernels for Section~\ref{sec:experiments-whitened}, in Table~\ref{tab:whitened-time-mat12}- \ref{tab:whitened-time-sqexp}. 
These results are consistent to our conclusion in the paper.
\begin{table}[!h]
  \centering
  \scalebox{.8}{\input{figures/whitening/Mat12-wallclock.tex}}
  \caption{Whitening time comparison (second) of HIP-GP v.s. SVGP with Matérn($0.5$) kernel.}
  \label{tab:whitened-time-mat12}
\end{table}

\begin{table}[!h]
  \centering
  \scalebox{.8}{\input{figures/whitening/Mat32-wallclock.tex}}
  \caption{Whitening time comparison (second) of HIP-GP v.s. SVGP with Matérn($1.5$) kernel.}
  \label{tab:whitened-time-mat32}
\end{table}

\begin{table}[!h]
  \centering
  \scalebox{.8}{\input{figures/whitening/SqExp-wallclock.tex}}
  \caption{Whitening time comparison (second) of HIP-GP v.s. SVGP with squared exponential kernel.}
  \label{tab:whitened-time-sqexp}
\end{table}

\vspace{5cm}



\subsection{UCI benchmark dataset}

We include another experiment on the UCI 3D Road dataset ($N= 278,319, D=3$).  Following the same setup as Wang et al., 2019, we train HIP-GP with $M= 36{,}000$ and a mean-field variational family, and compare to their reported results of Exact GP, SGPR ($M= 512$) and SVGP ($M= 1,024$) (Table~\ref{uci}). With the large $M$, HIP-GP achieves the smallest NLL, and the second-smallest RMSE (only beaten by exact GPs).

\begin{table}[!ht]
\centering
\scalebox{.8}{\input{figures/appendix/3droad-result}}
\caption{UCI 3D Road experiment ($N=278{,}319$). Results are averaged over 3 random splits.}
\label{uci}
\end{table}

%% file: figures/whitening/Mat12-wallclock.tex
\begin{tabular}{lrrrrr}
\toprule
$M$ &     $10^3$       & $10^4$     & $10^5$     & $10^6$  \\
\midrule
HIP-GP   &  $\mathbf{0.0078}$  &  $\mathbf{0.0187}$  &  $\mathbf{0.3484}$  &  $\mathbf{1.4727}$  \\
SVGP     &  $0.0152$ &   $0.1516$  &   n/a      & n/a       \\ 
\bottomrule
\end{tabular}

%% file: figures/whitening/Mat32-wallclock.tex
\begin{tabular}{lrrrrr}
\toprule
$M$ &     $10^3$       & $10^4$     & $10^5$     & $10^6$  \\
\midrule
HIP-GP   &  $\mathbf{0.0087}$  &  $\mathbf{0.0192}$  &  $\mathbf{0.3479}$  &  $\mathbf{1.4656}$  \\
SVGP     &  $0.0142$ &   $0.1379$  &   n/a      & n/a       \\
\bottomrule
\end{tabular}

%% file: figures/whitening/SqExp-wallclock.tex
\begin{tabular}{lrrrrr}
\toprule
$M$ &     $10^3$       & $10^4$     & $10^5$     & $10^6$  \\
\midrule
HIP-GP   &  $\mathbf{0.0112}$  &  $\mathbf{0.0199}$  &  $\mathbf{0.3683}$  &  $\mathbf{2.3433}$  \\
SVGP     &  $0.7090$ &   $0.0992$  &   n/a      & n/a       \\
\bottomrule
\end{tabular}

%% file: figures/appendix/3droad-result.tex
\begin{tabular}{llllllll}
    \toprule
     \multicolumn{4}{c}{\textbf{RMSE}}  & \multicolumn{4}{c}{\textbf{NLL}}\\
    \cmidrule(r){1-4} \cmidrule{5-8} 
    HIP-GP  & Exact GP & SGPR  & SVGP  &  HIP-GP  & Exact GP & SGPR  & SVGP \\
    \midrule 
    $0.189 $ & $\mathbf{0.101}$ & $0.661 $ & 
    $0.481$ & $\mathbf{-0.171} $ & $0.909 $ &$0.943$ & $0.697$ \\
    \bottomrule
    \end{tabular}